\documentclass{article}

\usepackage{arxiv}
\usepackage[final,nopatch=footnote]{microtype}
\usepackage[utf8]{inputenc} 
\usepackage[T1]{fontenc}    
\usepackage{hyperref}       
\usepackage{url}            
\usepackage{booktabs}       
\usepackage{amsfonts}       
\usepackage{nicefrac}       
\usepackage{microtype}      
\usepackage{lipsum}		
\usepackage{graphicx}
\graphicspath{{images/}}
\usepackage[numbers]{natbib}
\usepackage{eccvabbrv}
\usepackage{doi}
\usepackage{amsmath}
\usepackage{graphicx}
\usepackage{amsmath}
\usepackage{orcidlink}
\usepackage{graphicx}
\usepackage{booktabs}
\usepackage{subcaption}
\usepackage{rotating}
\usepackage{makecell}
\usepackage{graphicx}
\usepackage{booktabs}
\usepackage{siunitx}
\usepackage{geometry}
\usepackage{rotate}
\usepackage{longtable}
\usepackage{booktabs}
\usepackage{array}
\usepackage{geometry}
\usepackage{pdflscape}
\usepackage{longtable}
\usepackage{array}
\usepackage{booktabs}
\usepackage{longtable}
\usepackage{booktabs}
\usepackage{tabularx}
\usepackage{float}
\usepackage{setspace}
\usepackage{cleveref}
\usepackage{comment}

\usepackage[accsupp]{axessibility} 

\title{Comparative Analysis of YOLOv9, YOLOv10 and RT-DETR for Real-Time Weed Detection}

\author{%
\begin{minipage}[t]{0.45\textwidth}
    \centering
    {\fontsize{10}{12}\selectfont
    \href{https://orcid.org/0000-0002-8873-1638}{\includegraphics[scale=0.06]{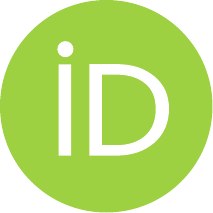}\hspace{1mm} \textbf{Ahmet O\u{g}uz Salt{\i}k}} \\
    \textnormal{Department of Artificial Intelligence in \\
    Agricultural Engineering \&} \\
    \textnormal{Computational Science Hub} \\
    \textnormal{University of Hohenheim} \\
    \texttt{ahmet.saltik@uni-hohenheim.de} \\
    }
\end{minipage}
\hspace{8mm} 
\begin{minipage}[t]{0.45\textwidth}
    \centering
    {\fontsize{10}{12}\selectfont
    \href{https://orcid.org/0009-0009-9206-5349}{\includegraphics[scale=0.06]{orcid.pdf}\hspace{1mm} \textbf{Alicia Allmendinger}} \\
    \textnormal{Department of Weed Science} \\
    \textnormal{University of Hohenheim} \\
    \texttt{alicia.allmendinger@uni-hohenheim.de} \\
    }
\end{minipage}
\\[7em]
\hspace{0mm} 
\begin{minipage}[t]{0.45\textwidth}
    \centering
    {\fontsize{10}{12}\selectfont
    \href{https://orcid.org/0000-0002-1808-9758}{\includegraphics[scale=0.06]{orcid.pdf}\hspace{1mm} \textbf{Anthony Stein}} \\
    \textnormal{Department of Artificial Intelligence in \\ Agricultural Engineering \&} \\
    \textnormal{Computational Science Hub} \\
    \textnormal{University of Hohenheim} \\
    \texttt{anthony.stein@uni-hohenheim.de} \\
    }
\end{minipage}
}

\renewcommand{\shorttitle}{Comparative Analysis of YOLOv9, YOLOv10 and RT-DETR for Real-Time Weed Detection}
\begin{document}
\maketitle
\renewcommand\thefootnote{}
\footnotetext{This preprint is a revised version of the paper submitted to the CVPPA Workshop at ECCV 2024.}
\renewcommand\thefootnote{\arabic{footnote}}
\begin{abstract}
This paper presents a comprehensive evaluation of state-of-the-art object detection models, including YOLOv9, YOLOv10, and RT-DETR, for the task of weed detection in smart-spraying applications focusing on three classes: Sugarbeet, Monocot, and Dicot. The performance of these models is compared based on mean Average Precision (mAP) scores and inference times on different GPU and CPU devices. We consider various model variations, such as nano, small, medium, large alongside different image resolutions (320px, 480px, 640px, 800px, 960px). The results highlight the trade-offs between inference time and detection accuracy, providing valuable insights for selecting the most suitable model for real-time weed detection. This study aims to guide the development of efficient and effective smart spraying systems, enhancing agricultural productivity through precise weed management.
\end{abstract}

\section{Introduction}
\label{sec:intro}

There is a pressing need to reduce the use of pesticides in agriculture, driven by policy initiatives such as the EU Green Deal and its Farm-to-Fork strategy \cite{fetting2020european}, as well as social interest. One strategy for reducing the amount of pesticides used is to decrease the use of herbicides. The predominant methodology for the application of herbicides is a broadcast application, in which herbicides are distributed uniformly throughout the field \cite{allmendinger2022precision}. However, an examination of the distribution and prevalence of weeds and weed species reveals a notable heterogeneity inside agricultural fields \cite {marshall1988field}. Weeds frequently manifest in aggregated patches of varying dimensions and composition \cite {gerhards2003real}. Consequently, when herbicides are applied uniformly across a field, the composition of weeds can remain consistent for an extended period \cite {gerhards2022advances}. However, the broadcast application of herbicides also results in their application in areas where there are no weeds.

Therefore, it would be advantageous to determine the location and identity of the weed species within a field prior to the application of an herbicide. On the one hand, areas where no weeds are present could be excluded from the treatment. On the other hand, specific products could be applied to the weeds present to achieve a higher degree of effectiveness \cite{allmendinger2022precision}. This information must be collected expeditiously in order to enable the application of an herbicide at the appropriate time and to avoid the necessity of a comprehensive and meticulously compiled assemblage of information. This information can be collected in real-time or offline. Real-time methods are continuous methods that comprise sensors mounted on the front of the tractor such as cameras that acquire images. These images are processed by an image classifier located, for example, in the tractor cabin \cite{allmendinger2022precision}. In the event that there are weeds, the information can be transmitted to the sprayer, which will then open the corresponding nozzle at the appropriate location. The term offline methods refers to discontinuous methods, which involves the use of a system, such as a drone, to acquire images of the field in advance \cite{mink2018multi}. The drones are usually provided by external service providers. The images are also subjected to an image classifier, although this need does not occur in the field \cite {mink2018multi}. An application map is generated, which can subsequently be loaded onto the terminal of the tractor and utilised for spraying purposes. This requires the use of RTK-GNSS \cite{mink2018multi}. In both online and offline contexts, the two methods are based on the same underlying principles. They rely on the processing of images with an algorithmic approach to enable differentiation between crops and weeds \cite{allmendinger2022precision}. 

In recent years, significant advances have been made in the field of crop and weed classification, some studies even focusing on the distinction between individual crop and weed species. There are various systems to differentiate between crops and weeds. These systems are based on machine learning, with a particular focus on deep learning, which encompasses two main areas: image classification and object detection. In image classification, the whole image is assigned to a category. In object detection, the coordinates of the object within the image are also detected \cite{7112511}. Object detection can be carried out with a two-stage detector or a one-stage detector \cite {liu2020deep}. Two-stage detectors include Fast R-CNN \cite{fast-r-cnn} and Faster R-CNN \cite{faster-r-cnn}. In the first step, a region proposal network is used to generate a limited set of regions of interest. In the second step, Convolutional Neural Networks (CNNs) \cite{lecun2015deep} are used to encode the extracted features and assign them to previously defined classes, as well as to predict the bounding boxes \cite{diwan2023object}. One-stage detectors include the You Only Look Once (YOLO) models started with the work of Redmon \etal \cite{YOLO1}. In a single step, the bounding boxes are predicted and assigned to previously defined classes. This approach enables real-time processing due to the inclusion of all steps in a single stage \cite{diwan2023object}. Several studies, which will be detailed in \Cref{sec:Related_work}, have already tested the use of one- and two-stage detectors for the classification of weeds. 

In recent years, there has been a further development of the use of neural networks in addition to the one- or two-stage detectors. In 2017, the use of transformers for natural language processing was introduced \cite {vaswani2017attention}. Transformers have the potential to become state-of-the-art deep learning models and outperform current models due to their self-attention mechanism that allows them to focus on several sequences in parallel \cite{vaswani2017attention}. Following the success of the Transformer in natural language processing, the application was extended to image classification. The objective was to divide the images into patches that are used as tokens for the transformer\cite{dosovitskiy2020image}. These vision-based transformers (ViT) have demonstrated the ability to outperform state-of-the-art models in image classification on a number of datasets \cite{dosovitskiy2020image}.
In the event that transformers are employed for the purpose of object detection, detection transformers (DETR) \cite{detr} are utilized. DETRs typically use a CNN such as ResNet as a backbone to extract feature maps \cite{carion2020end}. This enhances the spatial hierarchy and feature extraction capabilities. Furthermore, position encodings are incorporated to retain spatial information \cite{carion2020end}.  
To date, the use of transformers in the context of agricultural images has been largely confined to the domain of image classification with ViT \cite{sharma2023self}. Nevertheless, an increasing number of studies utilising transformers for object detection in an agricultural context will emerge as their robust performance enables the actual situation in an agricultural field to be processed \cite{li2024detr}.

Numerous studies in the agricultural context have focused on the comparison between one-stage and two-stage detectors, such as YOLO and Faster R-CNN. The utilisation of transformers in this context has been less extensively tested. In particular, the comparison of different state-of-the-art models for real-time application with RT-DETR \cite{zhao2024detrs}, a transformer-based object detection model that can be used for real-time application, needs to be further investigated. The objective of this study is to compare the approach of the YOLO one-stage detector, among all state-of-the-art models v9 \cite{yolov9} and v10 \cite{yolov10}, with the RT-DETR. The approaches will be tested as a real-time application in the context of a real agricultural field situation. For this purpose, images from a sugar beet field are used in which monocotyledonous and dicotyledonous weeds can be found beside the crop. In addition to the different models tested, the image size is also to be varied and compared in terms of the mAP score and the inference time of different GPUs. This comparison will enable a statement to be made about which models enable efficient and effective weed management, and thus increase the productivity of precise weed control methods.

\section{Related work}
\label{sec:Related_work}

\subsection{One-stage and two-stage approaches in weed detection}

Due to their precise and rapid detection capabilities, Rahman \etal compared the use of different one- and two-stage detectors \cite{rahman2023performance}. The data set comprised 848 images from three classes collected in cotton fields under different conditions. The models tested included YOLOv5, RetinaNet, as one-stage detector, and Fast and Faster RCNN, as two-stage detectors. RetinaNet achieved the highest mean average precision (mAP) mAP50-95 score of 62.97 \% among the one-stage detectors. The Faster R-CNN X101-FPN model achieved an mAP50-95 score of 61.48\% \cite{rahman2023performance}.
Saleem et al. analysed 17.509 images from eight classes in the DeepWeeds dataset \cite{saleem2022weed}. They compared different one-stage detectors, such as YOLOv4 and RetinaNet, and two-stage detectors, such as Faster Region-based Convolutional Neural Network and Region-based Fully Convolutional Network. With the default settings, an mAP50-95 score of 79.68\% was achieved with Yolov4 and 87.64\% with Faster R-CNN Resnet-101. Furthermore, different methods were used to enhance the networks with the objective of improving the mAP score for detection and classification. With Faster R-CNN Resnet-101, an improvement of 5.8\% of the mAP50-95 score was achieved \cite{saleem2022weed}. 
Dang et al. compared in their study 25 state-of-the-art Yolo models using the CottonWeedDet12 cotton dataset, which consists of 5648 images and 12 weed classes. The mAP50-95 scores varied depending on the model, with the highest being 89.48\% for Yolov4 and the lowest 68.18\% for Yolov3-tiny \cite{dang2023yoloweeds}.

\subsection{Transformers in crop production settings}
The use of vision-based transformers has already been investigated in several studies. For example, Suma et al. employed in their study a combination of CentreNet and a ViT (CETR) for the purpose of wheat head detection. The CETR achieved a higher mAP50 score with 83.18 \% than the CNN, which achieved a score of 71.6 \% \cite{suma2024original}.
Zhao et al. implemented a Real-time Detector (RT-DETR), which is an extension of the DETR \cite{zhao2024detrs} algorithm designed for real-time object detection processing, achieving state-of-the-art performance. This surpassed the performance of previous YOLO models.  
In the context of the differentiation between crop and weed, RT-DETR was not used by now. There are some studies in the agriculture context for example the study, conducted by Aguilera et al. \cite{aguilera2023comprehensive}. The purpose was to classify the ripeness of blueberries. Given that blueberries typically grow in clusters, inaccuracies and imprecise bounding boxes are often observed during recognition. The use of RT-DETR resulted in the identification of a greater number of blueberries than that achieved with other models \cite{aguilera2023comprehensive}. Furthermore, the inference time of the RT-DETR-L with 11.551 ms was comparable to the inference time of the default YOLOv7 model with 8.059 ms \cite{aguilera2023comprehensive}.

In contrast to the studies briefly reviewed above, our work focuses on the comparison of the latest models YOLOv9, YOLOv10, and RT-DETR for use in weed detection. The models are tested on images from the agricultural context to determine their potential for real-time application.

\section{Material and Methods}
\label{sec:Material_and_Methods}

This section explains the methodologies applied in the study, encompassing data preparation, model selection, evaluation metrics, and experimental procedures. The following subsections provide an in-depth description of the dataset used for training and testing (\Cref{subsec:data_description}), the different versions of the YOLO (YOLOv9, YOLOv10) and RT-DETR models used (\Cref{subsec:Weed_Detection_Models}), the metrics used to evaluate model performance (\Cref{subsec:Evaluation_Metrics}), and the specific experimental setup (\Cref{subsec:Experimental_Setup}).

\subsection{Data Description}
\label{subsec:data_description}

In an ongoing research project in Rhineland-Palatinate, Germany, a comprehensive dataset of 2074 high-resolution images (1752 x 1064 pixels) was collected using industrial camera sensors mounted on a herbicide sprayer attached to a tractor moving at 1.5 m/s. Each field camera unit (FCU), equipped with a 2.3-megapixel RGB sensor, a 6mm effective focal length (EFL), and a dual-band filter for capturing RED and near-infrared (NIR) wavelengths, was  positioned on the sprayer’s linkages at a constant height of 1.1 meters and a 25-degree off-vertical angle. The images, capturing sugar beet crops along with dicot and monocot weeds, were taken in an outdoor experimental setup where crops and weeds were grown in boxes placed on Euro pallets with varying soil conditions to create a controlled environment. The machine, equipped with these camera sensors, navigated through these clearly labeled boxes to facilitate the precise differentiation of soil and weed types. Following data collection, pseudo-RGB images were generated from the raw RED and NIR wavelengths through projection correction, and experts manually labeled these images by focusing on the primary crop class Sugarbeet and two comprehensive weed classes: Dicot and Monocot \cite{Sourav_ARCS}. These two weed types were selected for annotation because commercial herbicide applications are formulated based on broad weed species rather than targeting specific weed species \cite{monocot_dicot}.

\begin{figure}[tb!]
  \centering
  \begin{minipage}{0.49\textwidth}
    \centering
    \includegraphics[width=\textwidth]{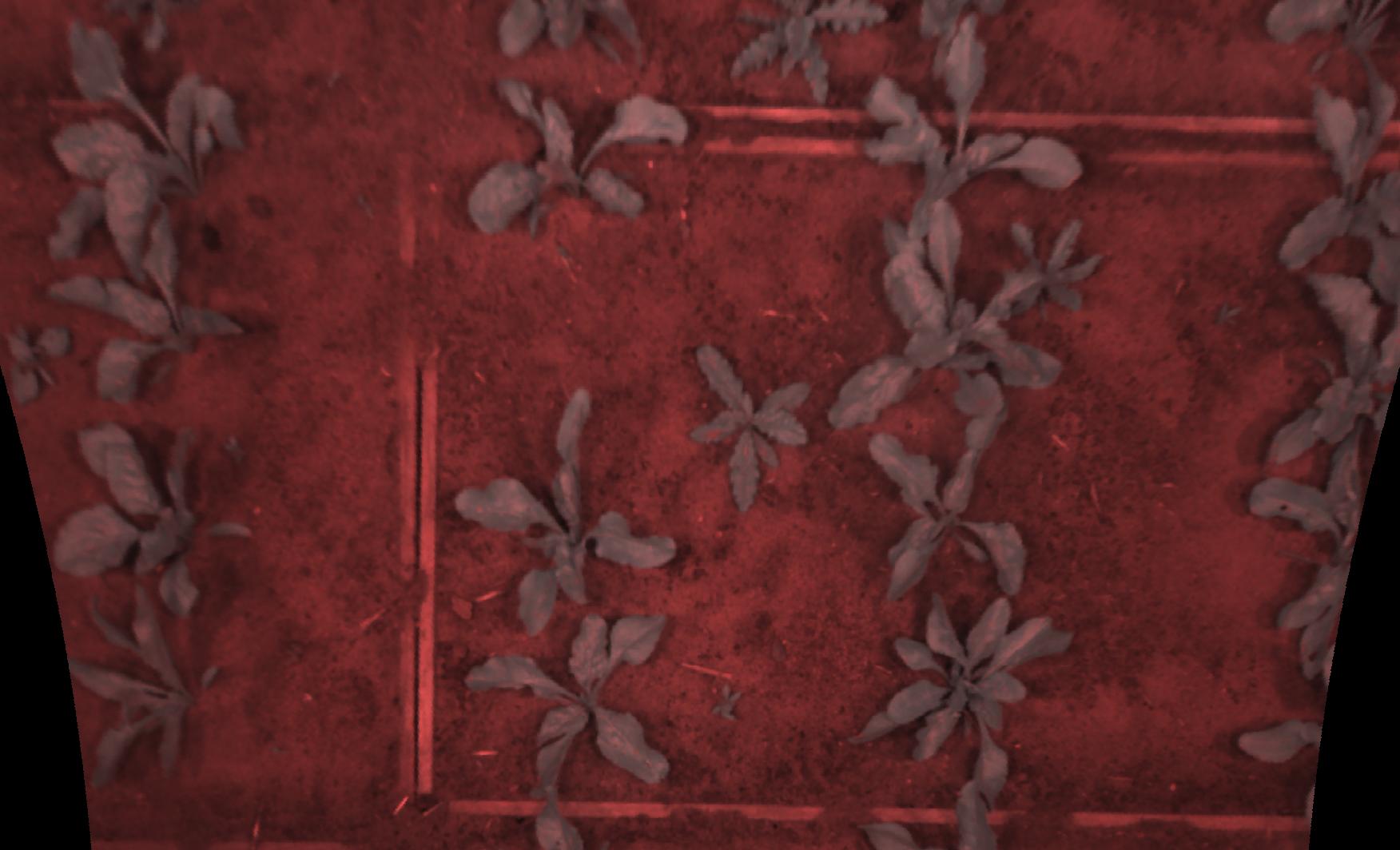}
  \end{minipage}
  \hfill
  \begin{minipage}{0.49\textwidth}
    \centering
    \includegraphics[width=\textwidth]{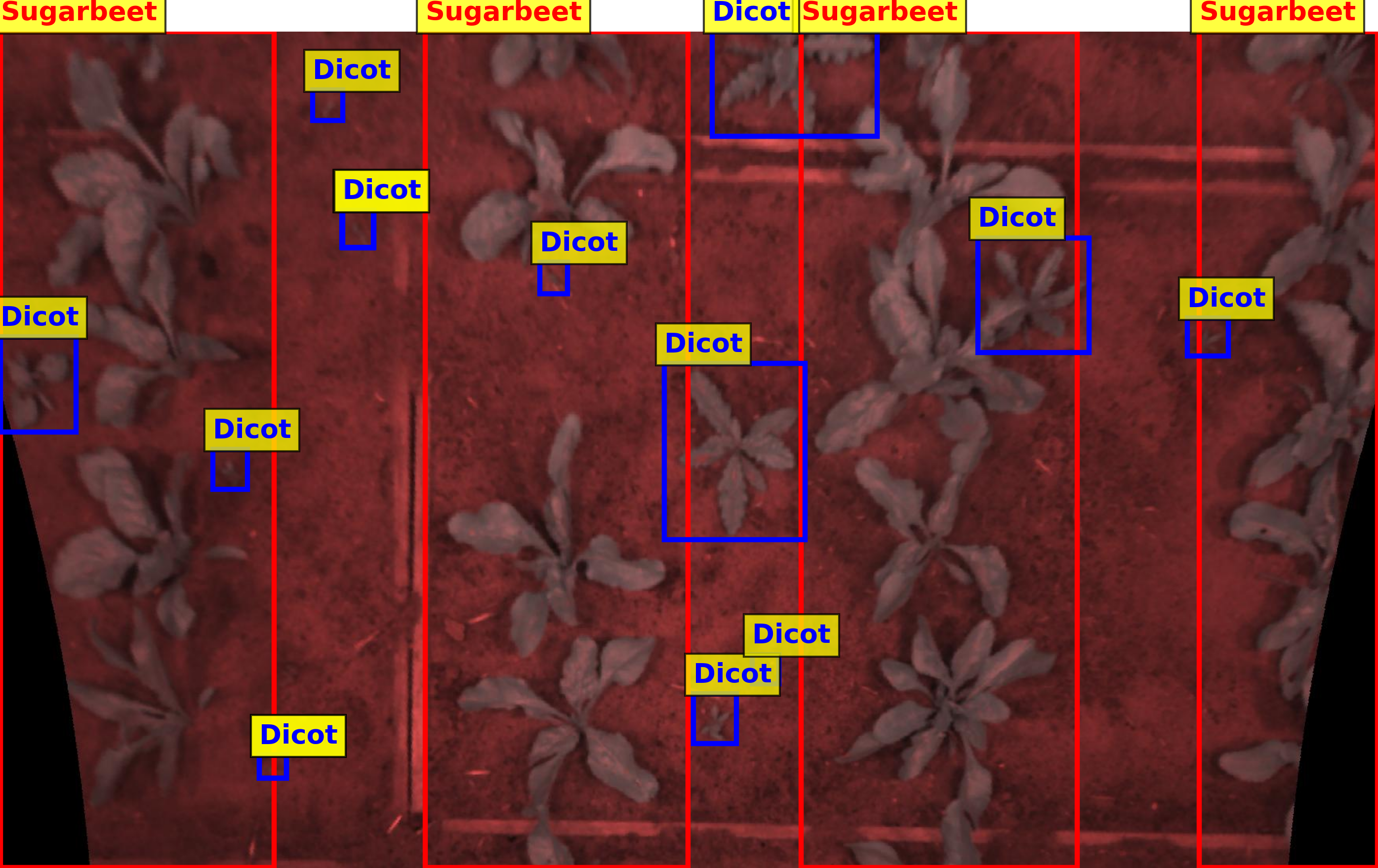}
  \end{minipage}

  \vspace{0.25cm} 

  \begin{minipage}{0.49\textwidth}
    \centering
    \includegraphics[width=\textwidth]{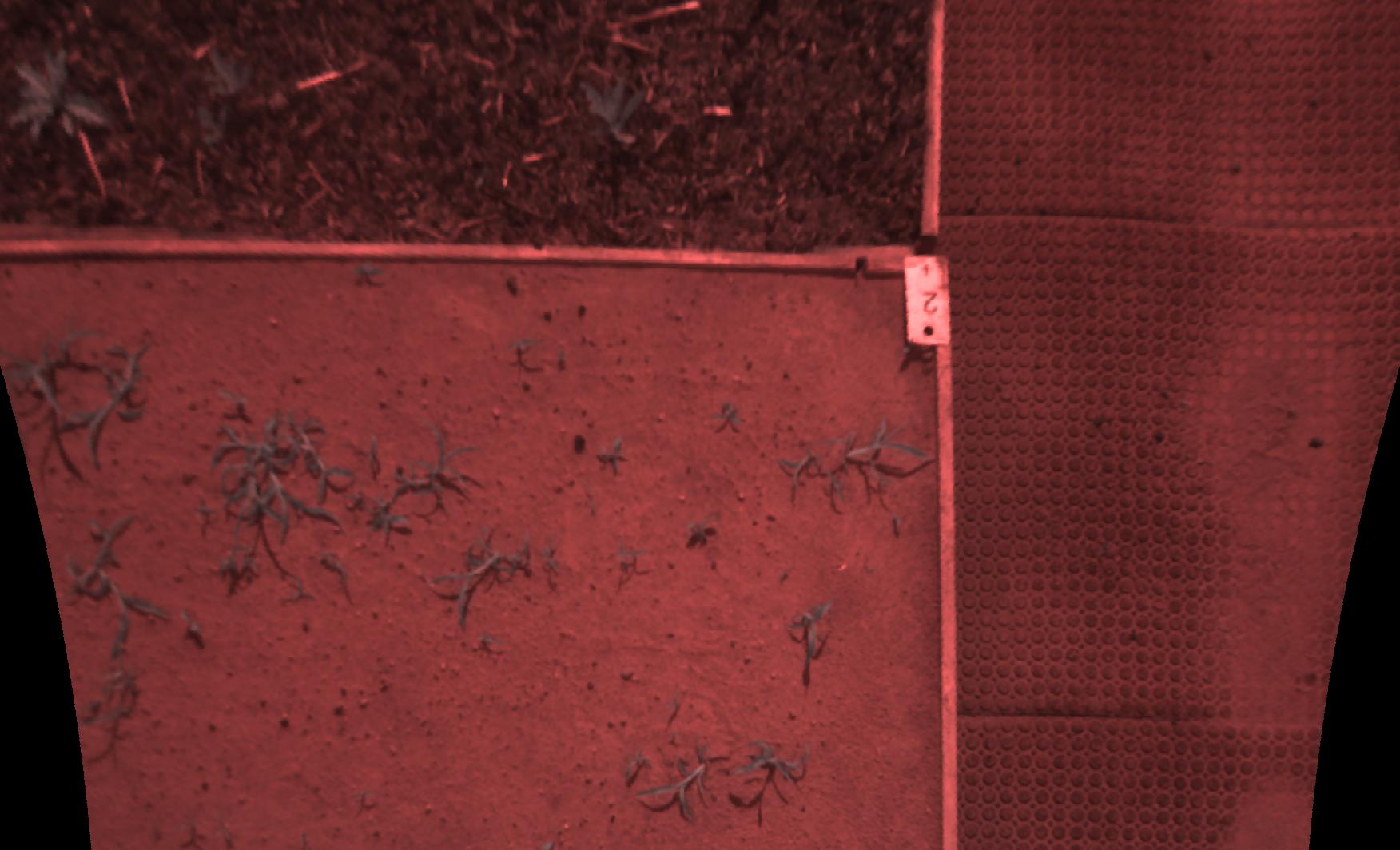}
    \caption*{\textbf{(a)} Captured images}
  \end{minipage}
  \hfill
  \begin{minipage}{0.49\textwidth}
    \centering
    \includegraphics[width=\textwidth]{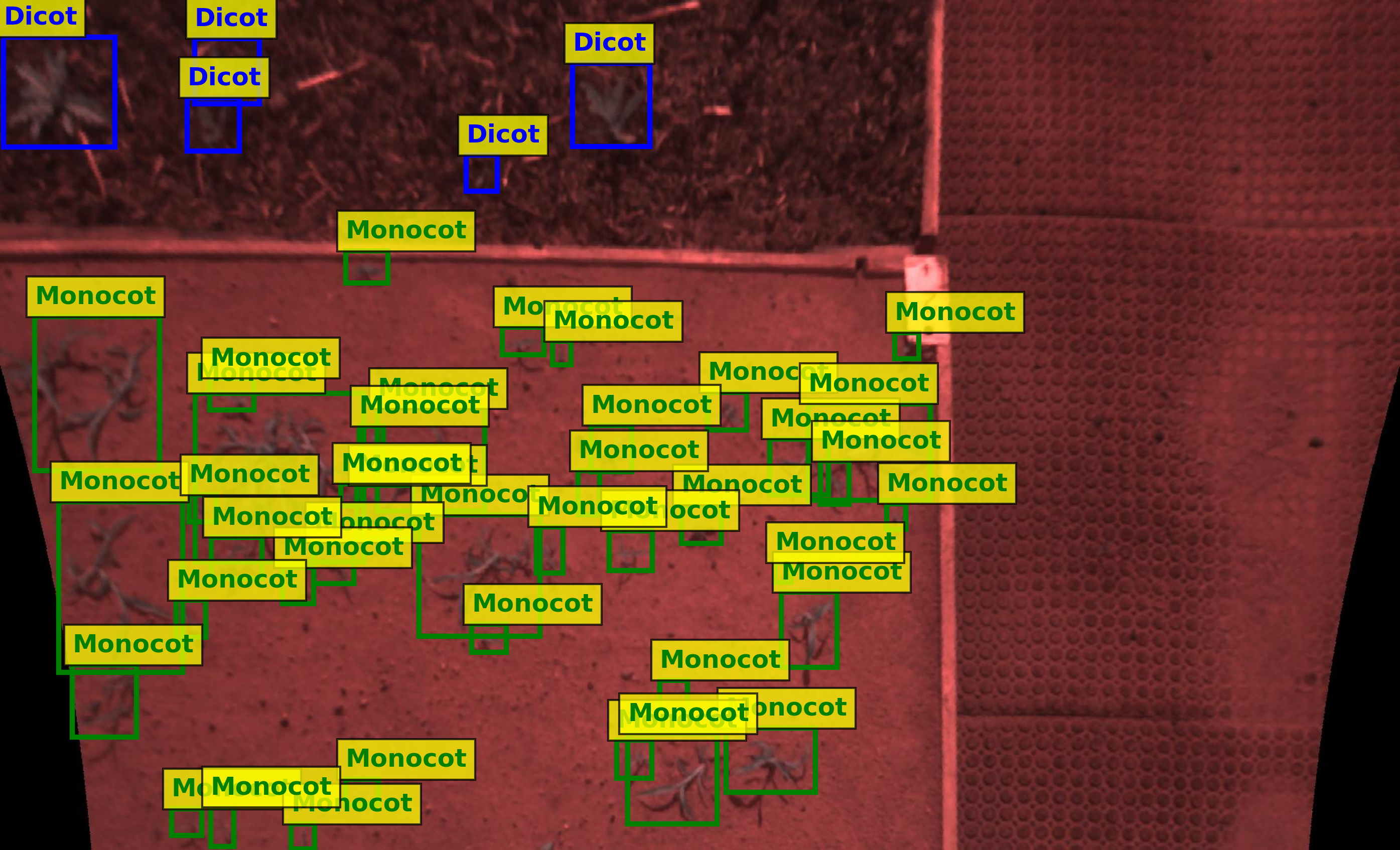}
    \caption*{\textbf{(b)} Annotated ground truth}
  \end{minipage}

  \caption{Comparison of captured and annotated ground truth images.}
  \label{fig:comparison_groundtruth_annotated}
\end{figure}

\subsection{Weed Detection Models}
\label{subsec:Weed_Detection_Models}

\subsubsection{YOLOv9}
\label{subsec:YOLOv9}

is an advanced model designed for real-time object detection, addressing information loss commonly seen in deep neural networks and incorporating several key innovations to improve performance. Firstly, the Programmable Gradient Information (PGI) mechanism prevents information bottlenecks, ensuring that crucial data is preserved across network layers. This leads to more reliable gradient generation and better model convergence, which is essential to maintain high accuracy in detection tasks. Secondly, the use of Reversible Functions allows the network to invert data without loss, maintaining the integrity of information throughout the network's depth. This is particularly advantageous for lightweight models that typically face significant data loss during processing. Lastly, the Generalized Efficient Layer Aggregation Network (GELAN) optimizes parameter usage and computational efficiency by allowing flexible integration of various computational blocks, ensuring YOLOv9 can adapt to a wide range of applications while maintaining speed and accuracy \cite{yolov9, ultralytics}.

\subsubsection{YOLOv10}
\label{subsec:YOLOv10}

builds on the advancements of previous YOLO models with a more efficient architecture that eliminates the need for non-maximum suppression (NMS) during training. Substantial improvements include a dual assignment strategy, which enhances performance by reducing computational overhead and improving model accuracy, allowing for more efficient processing and better utilization of resources. Furthermore, the optimized architecture of YOLOv10 is designed for speed and accuracy, with the YOLOv10-S and YOLOv10-X models being significantly faster and more efficient than comparable models such as RT-DETR-R18 and RT-DETR-R101, while still maintaining high accuracy levels. Benchmark performance tests have shown outstanding results on standard datasets such as COCO\footnote{\url{https://cocodataset.org/}}\cite{COCO}, with YOLOv10 variants showing reduced parameter counts and lower latency, making the model more efficient without compromising detection performance \cite{yolov10, ultralytics}.

\subsubsection{Realtime Detection Transformer (RT-DETR)}
\label{subsec:RT-DETR}

leverages transformer-based architectures to enhance object detection in real-time scenarios. Developed by Baidu\footnote{\url{https://ir.baidu.com/}}, RT-DETR offers several significant advantages. The inclusion of attention mechanisms improves the detection of objects in complex and varied scenes, making RT-DETR particularly effective for high-speed processing and accuracy. These mechanisms enable the model to focus on relevant parts of the image, enhancing detection performance. Additionally, the efficient design of RT-DETR models, such as RT-DETR-R18 and RT-DETR-R101, ensures competitive performance with reduced latency and computational requirements, which is beneficial for processing large-scale and high-resolution images. The transformer-based approach also allows RT-DETR to excel in scenarios involving intricate and diverse scenes, providing robust detection performance through its advanced architecture \cite{zhao2024detrs, ultralytics}.

\subsection{Evaluation Metrics}
\label{subsec:Evaluation_Metrics}

In this study, standard object detection metrics \cite{padilla2020survey} were used to evaluate the performance of the model. Notably, mAP50 and mAP50-95 are the most frequently used metrics in the weed detection literature, as discussed in \Cref{sec:Related_work}.


\subsubsection{Precision and Recall}
Precision and recall are fundamental metrics in evaluating the performance of object detection models.

\begin{itemize}
    \item \textbf{Precision} is the ratio of true positive detections to the total number of positive detections (true positives + false positives).
    \item \textbf{Recall} is the ratio of true positive detections to the total number of actual positives (true positives + false negatives).
\end{itemize}

\subsubsection{Intersection over Union (IoU)}
is a fundamental metric in object detection that measures the overlap between two bounding boxes: the predicted bounding box and the ground truth bounding box.

\subsubsection{Mean Average Precision at IoU 0.50 (mAP50)}
calculates the average precision (AP) at an IoU threshold of 0.50. Precision and recall are computed for different confidence levels, and the precision-recall curve is plotted. The average precision is derived from the area under this curve (AUC). 

\subsubsection{Mean Average Precision from IoU 0.50 to 0.95 (mAP50-95)}
extends the mAP50 metric by evaluating AP at multiple IoU thresholds ranging from 0.50 to 0.95, in increments of 0.05. This provides a more comprehensive assessment of the accuracy of the model.

\subsubsection{Inference Time}
is the duration it takes for a model to process an input and produce an output. It is a crucial performance metric, particularly for real-time applications. Factors such as model complexity, hardware specifications, and optimization techniques can impact inference time. Assessing inference time helps in understanding the trade-offs between model accuracy and speed, ensuring the model meets the necessary performance standards for practical deployment.

These metrics are essential for evaluating the performance of object detection models, with mAP50 offering a baseline accuracy measure, mAP50-95 providing a detailed assessment across varying localization precisions, and inference time ensuring the model's practicality in real-world applications.

\subsection{Experimental Setup}
\label{subsec:Experimental_Setup}

\begin{figure}[tb!]
  \centering
  \includegraphics[width=0.99\textwidth]{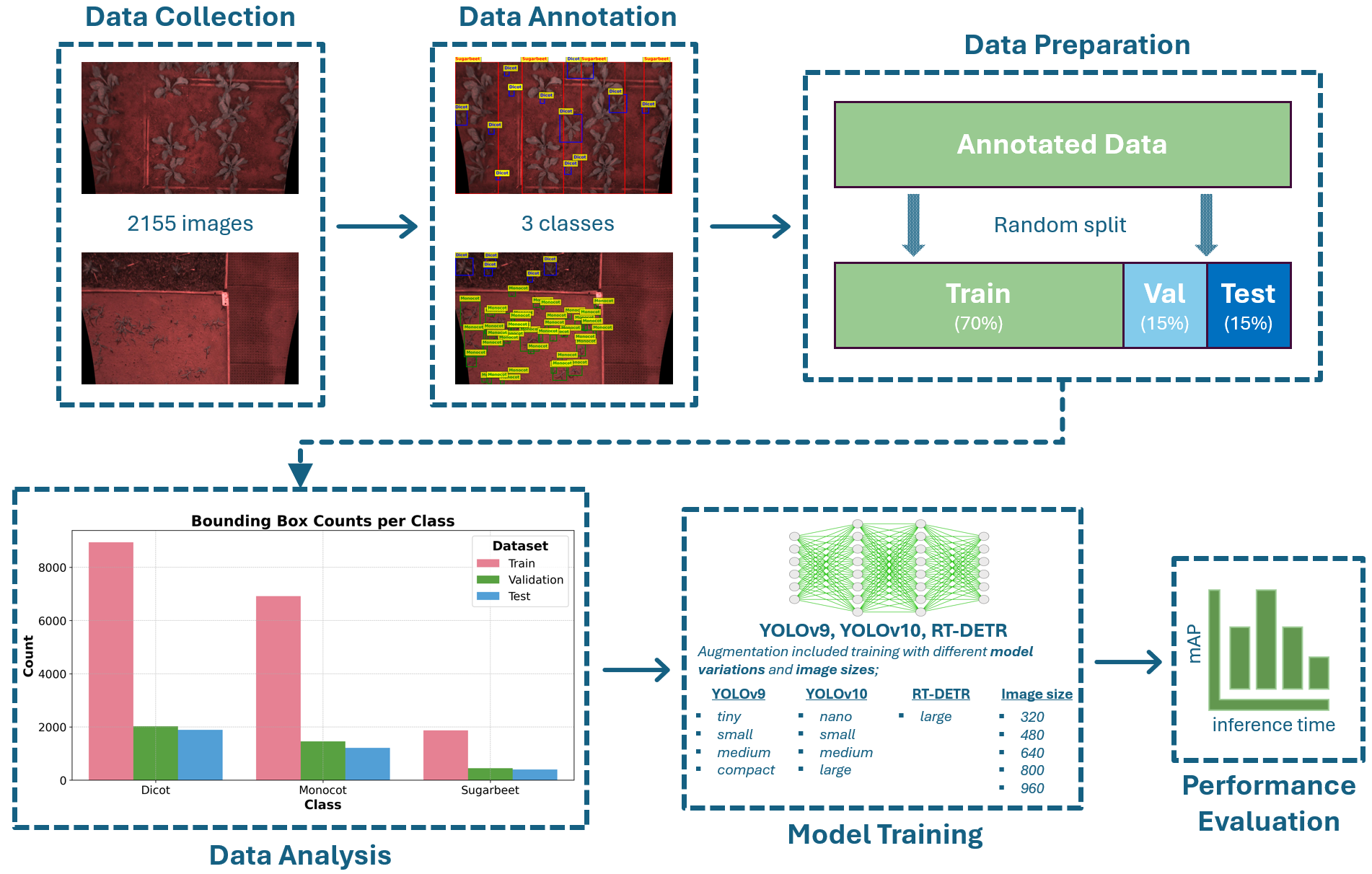}
  \caption{Machine Learning workflow as applied in this study.}
  \label{fig:pipeline}
\end{figure}

The experimental setup consists of several key stages: data collection, data annotation, data preparation, data analysis, model training, and performance evaluation. \Cref{fig:pipeline} illustrates our workflow.

A dataset of 2155 images was collected as described in \Cref{subsec:data_description}, focusing on the identification of three distinct classes: Dicot, Monocot, and Sugarbeet. The images were annotated manually to label the objects of interest within each image. The annotation process identified and labeled the objects in all images, resulting in bounding boxes for three classes. The annotated dataset was randomly split into three subsets: Training (70\%), validation (15\%), and testing (15\%). An analysis of the counts of the bounding boxes per class was performed for each subset.

Different versions of YOLO (YOLOv9, YOLOv10) and RT-DETR were employed for the object detection task using the Ultralytics framework \cite{ultralytics}. A range of available model sizes (tiny, nano, small, medium, large, compact) and image resolutions (320, 480, 640, 800, 960) were trained with activated online augmentation during the training process as defined in \cite{ultralytics}\footnote{\url{https://docs.ultralytics.com/modes/train/\#augmentation-settings-and-hyperparameters}} and using the COCO\footnote{\url{https://cocodataset.org/}}\cite{COCO} pretrained weights for fine-tuning. Models of the largest configurations were excluded from the implementation due to their substantial size, making them unsuitable for real-time scenarios where faster inference times and lower computational demands are critical \cite{Nils_Herterich}. The models were trained with PyTorch 2.3.1 and CUDA 12.1 \cite{pytorch} using a NVIDIA A100-SXM4-40GB GPU with 80GB RAM.

The training process consisted of 250 epochs, with early stopping implemented if the model's performance did not improve for 50 consecutive epochs. The learning rates were differentiated according to model type, set at 0.01 for YOLO models and 0.001 for RT-DETR models. This adjustment was made because selecting a learning rate of 0.01 for RT-DETR models resulted in observed fluctuations. Batch sizes were also adjusted according to the image resolution; a batch size of 16 was used for images up to 640 pixels, while a batch size of 8 was designated for images exceeding this resolution. The utilized optimizer was ADAMw, known for its decoupled weight decay regularization \cite{ADAMw}. In addition, a cosine learning rate scheduler was applied to dynamically adjust the learning rate during training. The hyperparameter setting can be seen in \Cref{tab:hyperparameters}.

Finally, the trained models were individually tested using the same image resolutions. Their evaluation was conducted using mean average precision (mAP50 and mAP50-95) and inference time metrics on two different NVIDIA GPU models: RTX3090, RTX4090, and Intel Core i9-14900K (32-core) CPU. To optimize real-time deployment efficiency, trained model files were converted to TensorRT\footnote{\url{https://developer.nvidia.com/tensorrt}} for GPU inferences and OpenVINO\footnote{\url{https://docs.openvino.ai/2024/index.html}} for Intel CPU inferences.

\begin{table}[tb!]
  \caption{Hyperparameters}
  \label{tab:hyperparameters}
  \centering
  \begin{tabular}{@{}ll@{}}
    \toprule
    \textbf{Hyperparameter} & \textbf{Value}\\
    \midrule
    Epoch & 250\\
    Patience & 50\\
    Learning rate & 0.01 for YOLOs \& 0.001 for RT-DETR\\
    Batch size & 16 for image size until 640px \& 8 for image size above 640px\\
    Optimizer & ADAMw (Decoupled Weight Decay Regularization)\\
    Learning rate scheduler & Cosine\\
    \bottomrule
  \end{tabular}
\end{table}

\section{Results}
\label{sec:results}

\begin{figure}[!]
  \centering
  \includegraphics[width=0.99\textwidth]{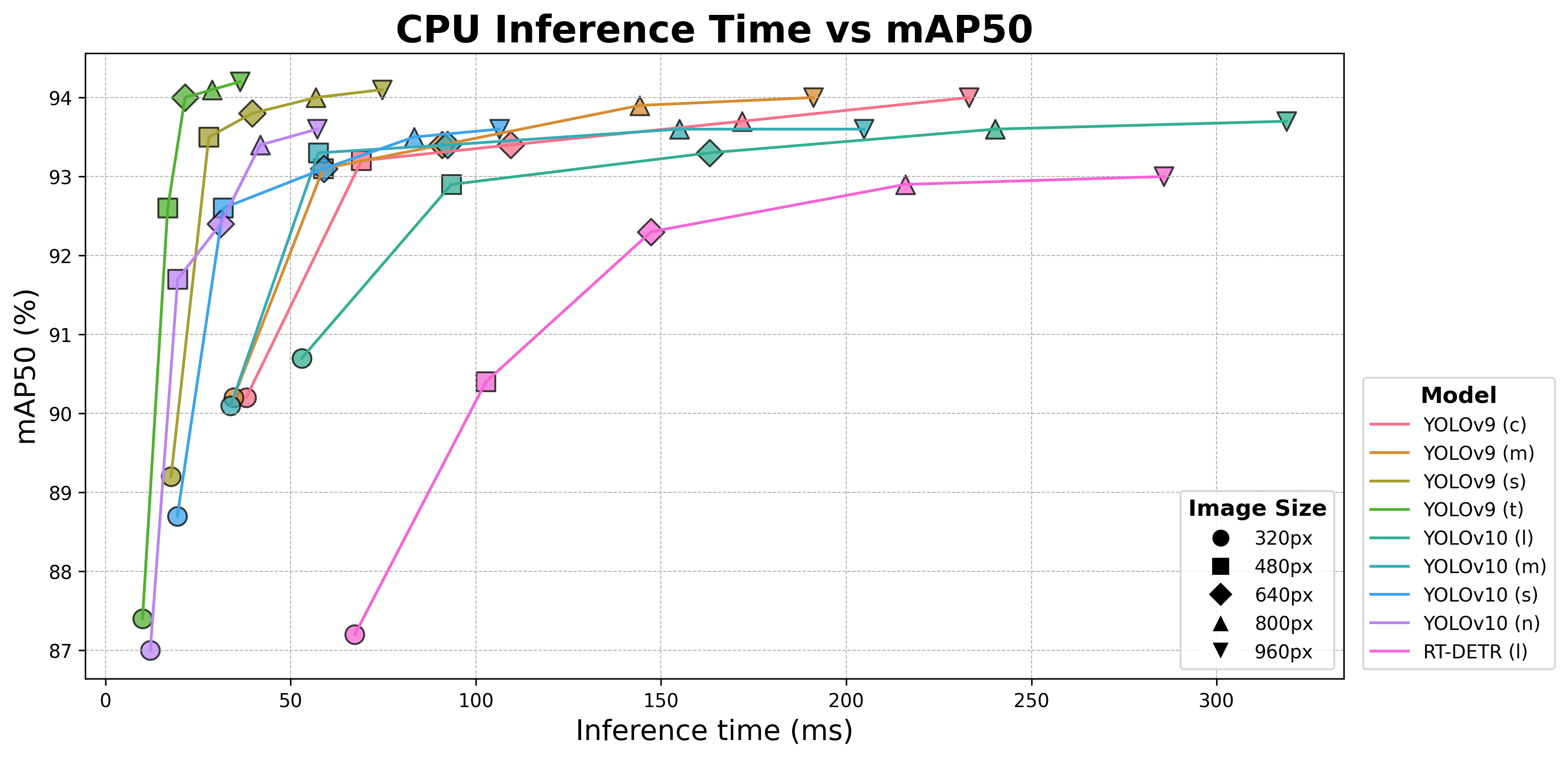}
  \caption{mAP50 vs inference time comparison on Intel Core i9-14900K (32-core) CPU.}
  \label{fig:map50}
\end{figure}

\begin{figure}[!]
  \centering
  \includegraphics[width=0.99\textwidth]{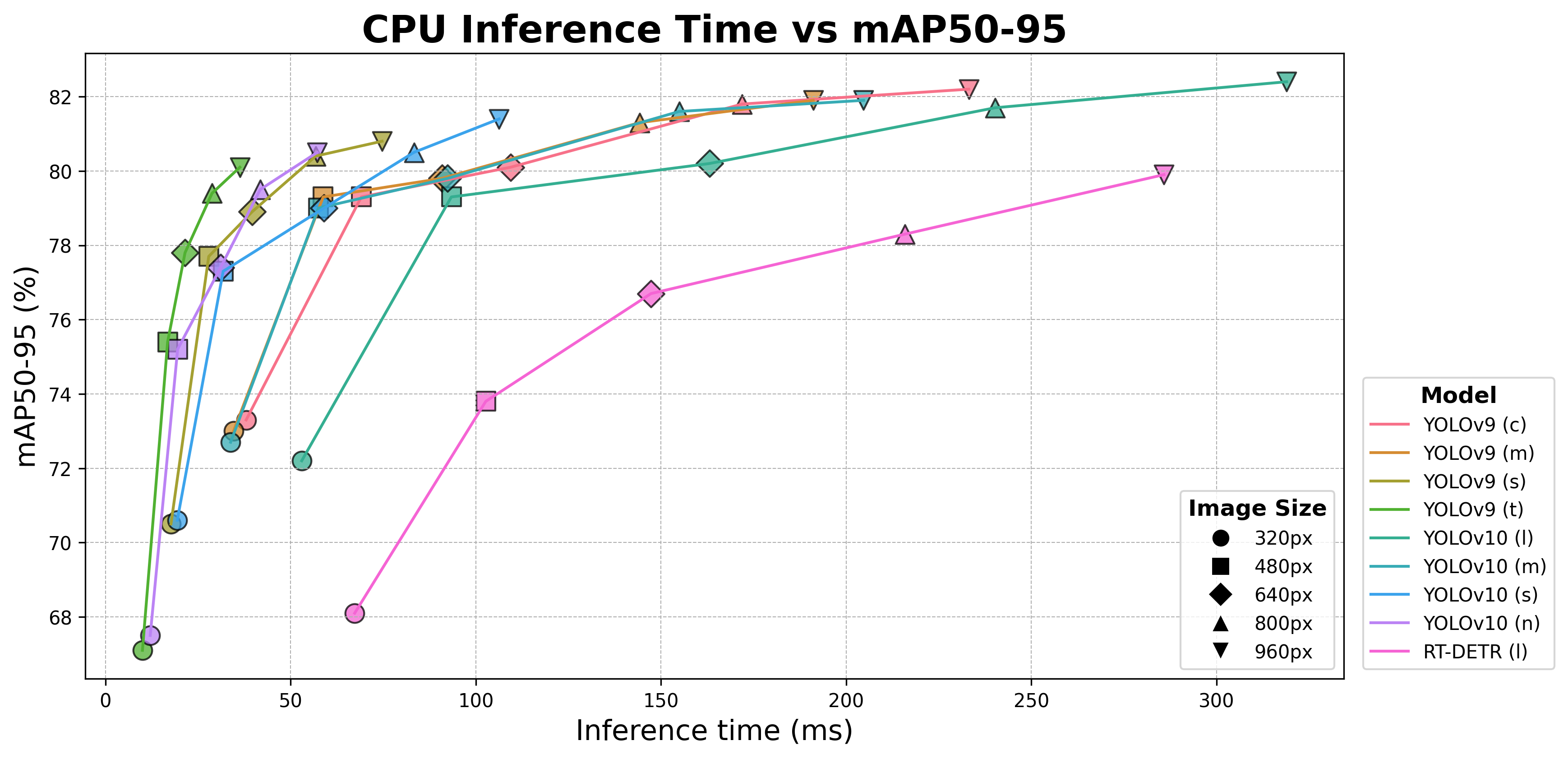}
  \caption{mAP50-95 vs inference time comparison on Intel Core i9-14900K (32-core) CPU.}
  \label{fig:map50-95}
\end{figure}

\begin{figure}[!]
  \centering
  \begin{subfigure}[b]{0.24\textwidth}
    \includegraphics[width=\textwidth]{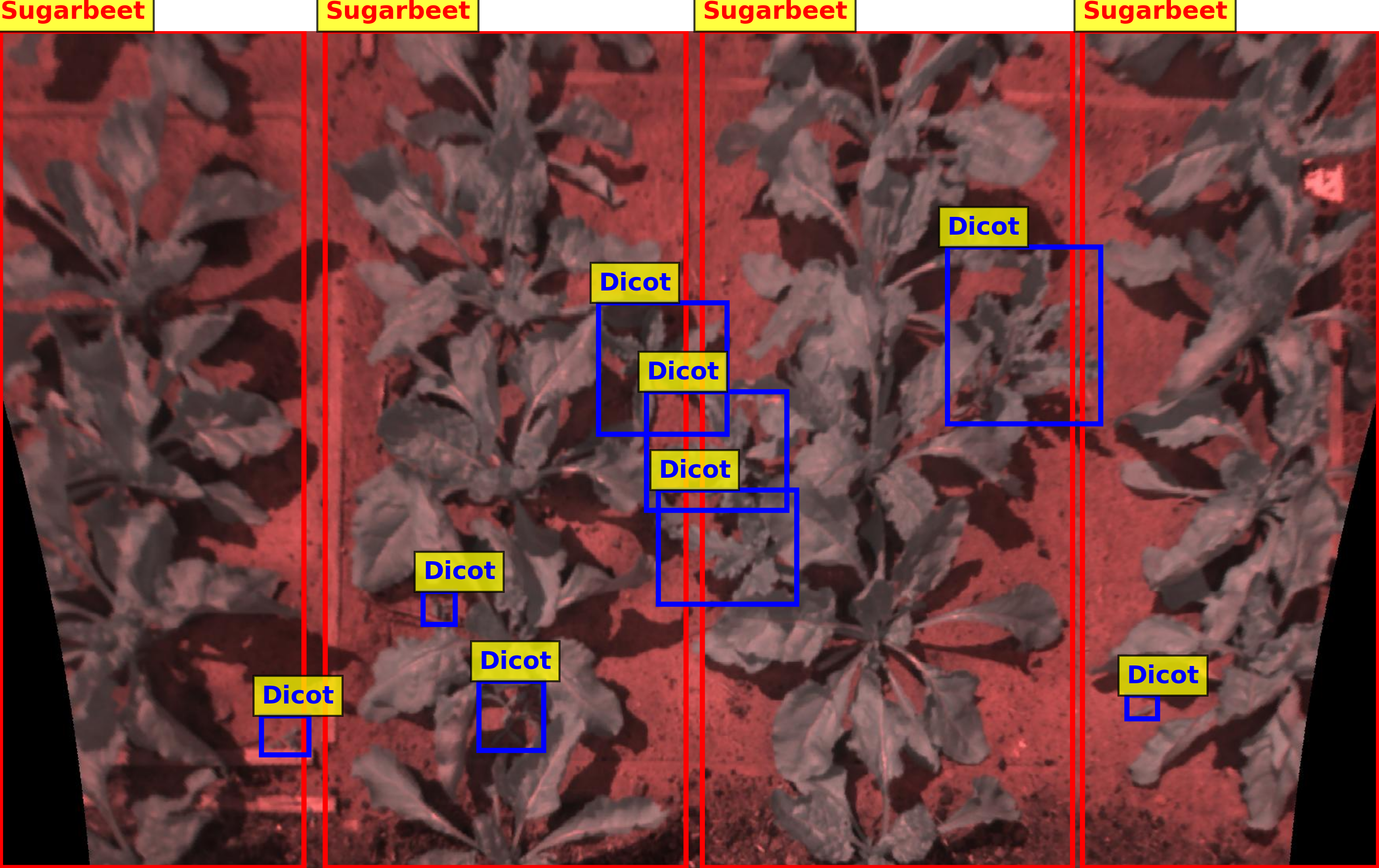}
    \caption{Ground truth}
    \label{fig:subimage1}
  \end{subfigure}
  \hfill
  \begin{subfigure}[b]{0.24\textwidth}
    \includegraphics[width=\textwidth]{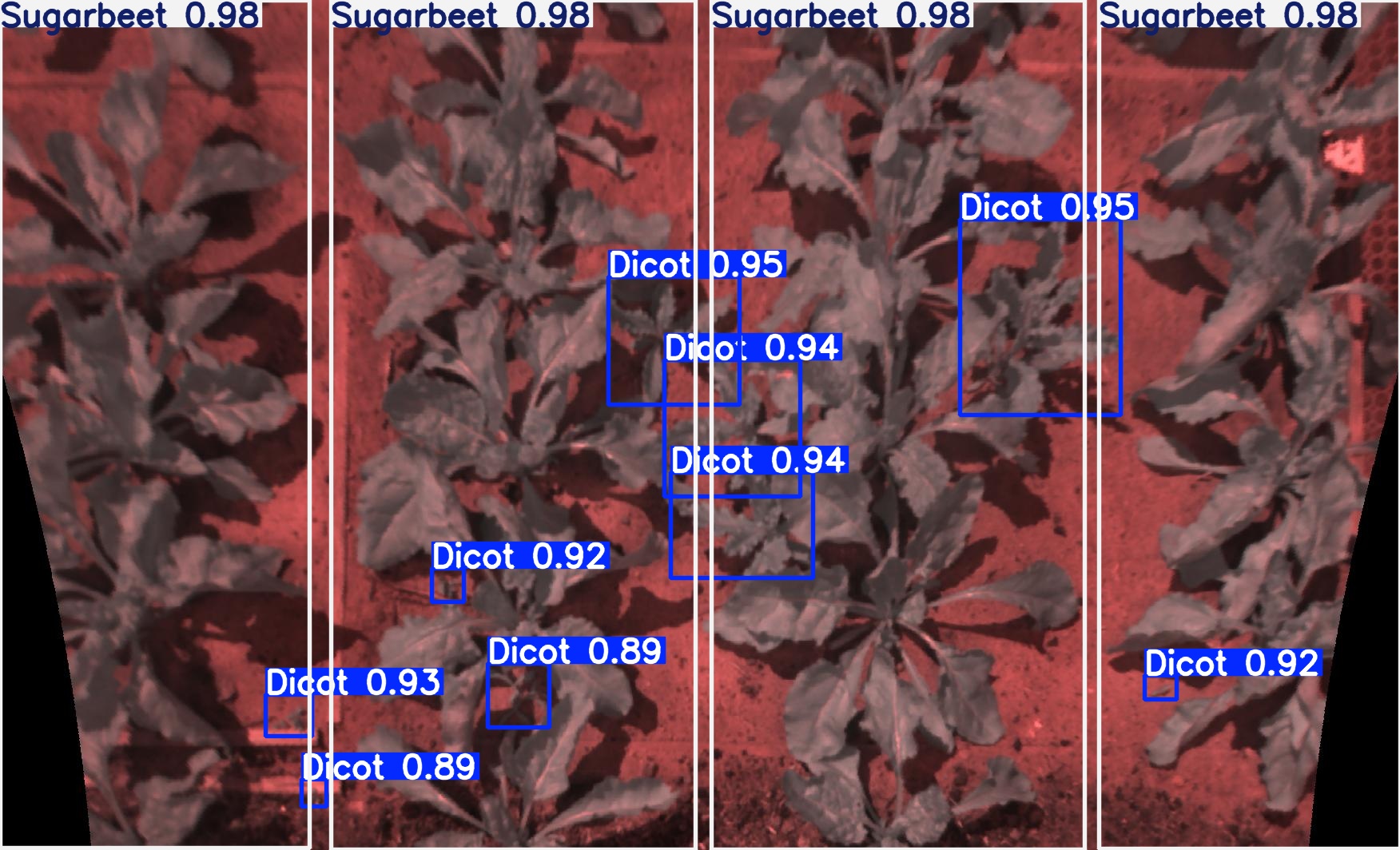}
    \caption{YOLOv9 (c)}
    \label{fig:subimage2}
  \end{subfigure}
  \hfill
  \begin{subfigure}[b]{0.24\textwidth}
    \includegraphics[width=\textwidth]{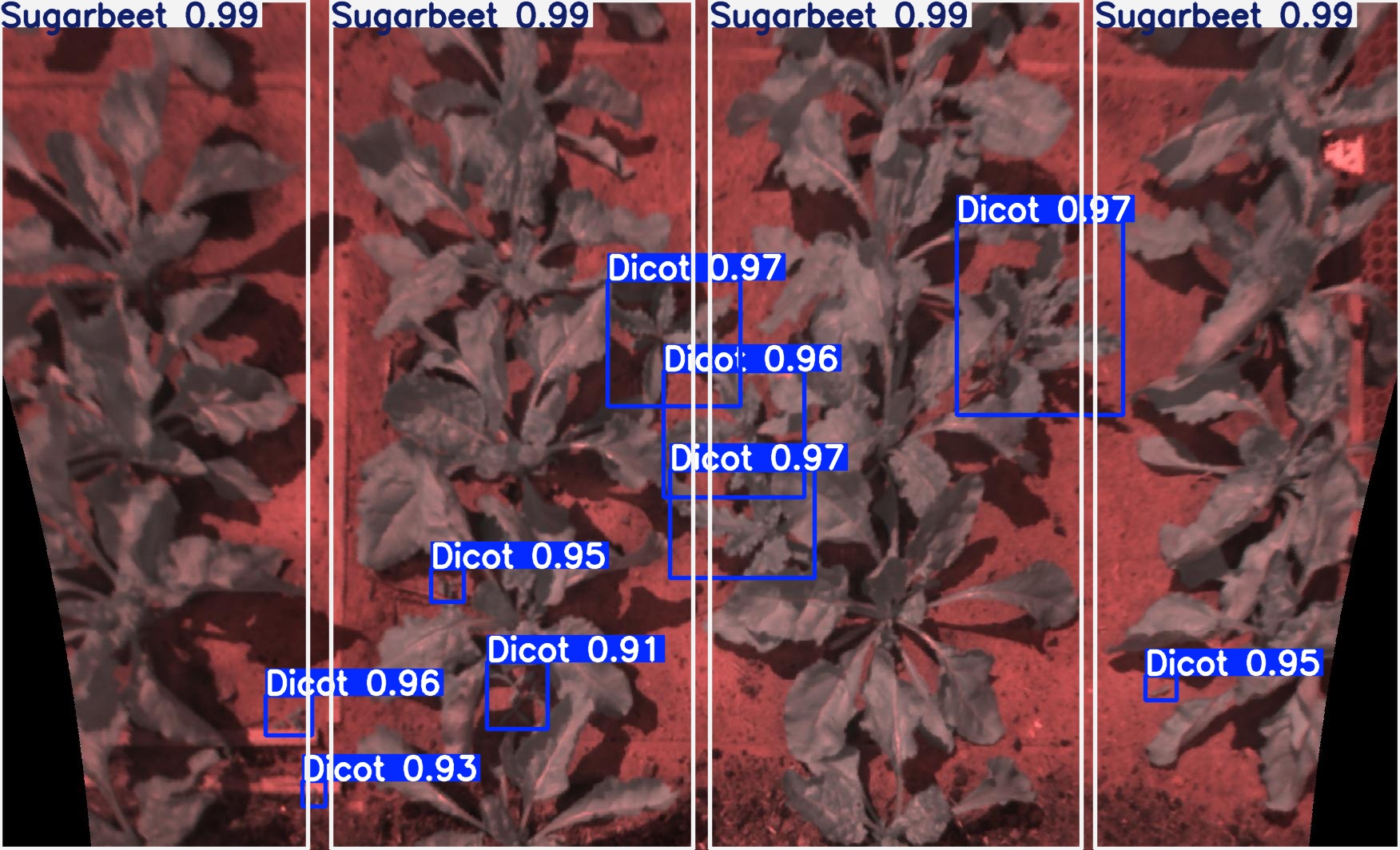}
    \caption{YOLOv10 (l)}
    \label{fig:subimage3}
  \end{subfigure}
  \hfill
  \begin{subfigure}[b]{0.24\textwidth}
    \includegraphics[width=\textwidth]{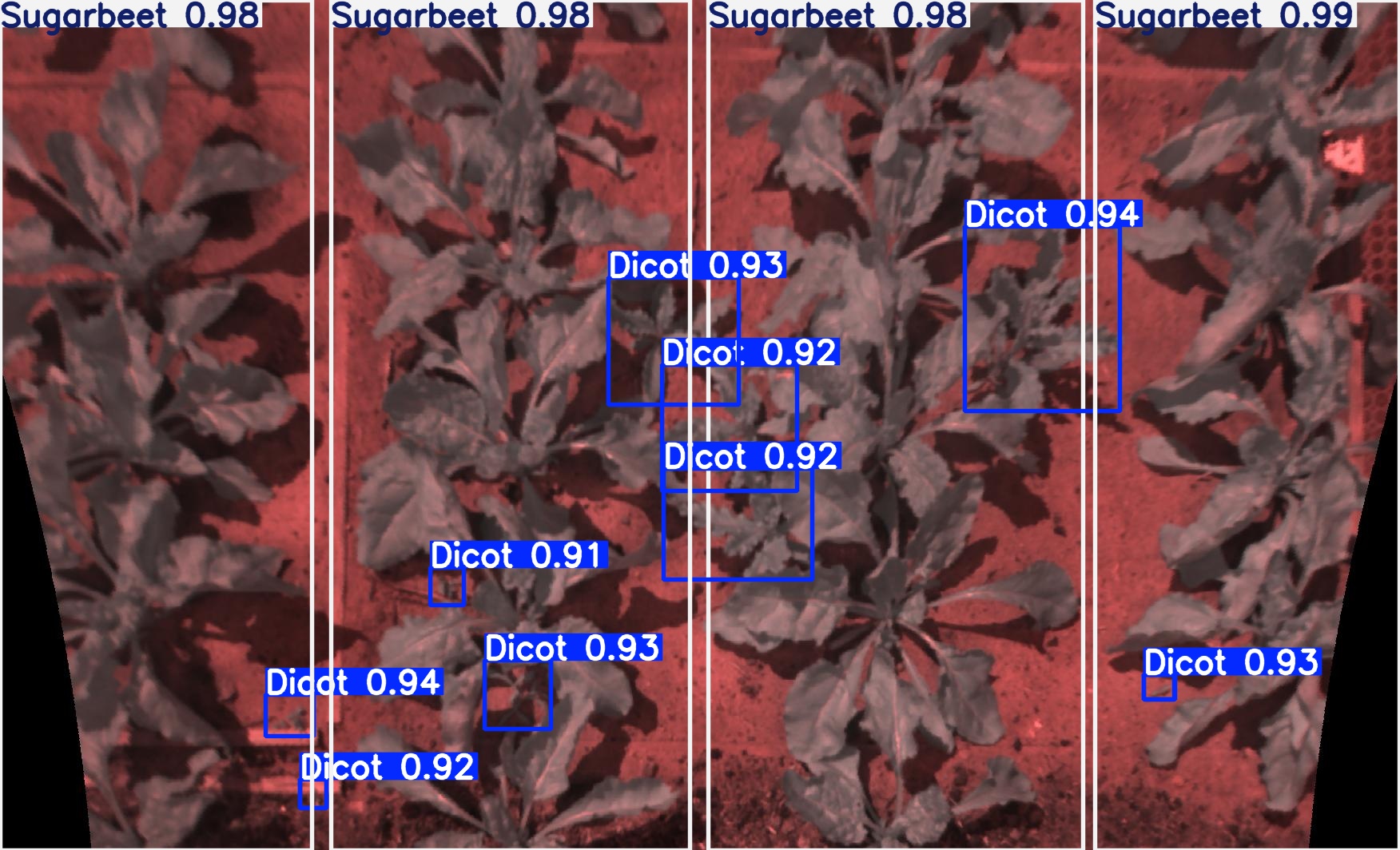}
    \caption{RT-DETR (l)}
    \label{fig:subimage4}
  \end{subfigure}
  \caption{Comparison of ground truth and prediction results from various models using 960-pixel image resolution.}
  \label{fig:new_predictions}
\end{figure}

\begin{figure}[!]
  \centering
  \begin{subfigure}[b]{0.32\textwidth}
    \includegraphics[width=\textwidth]{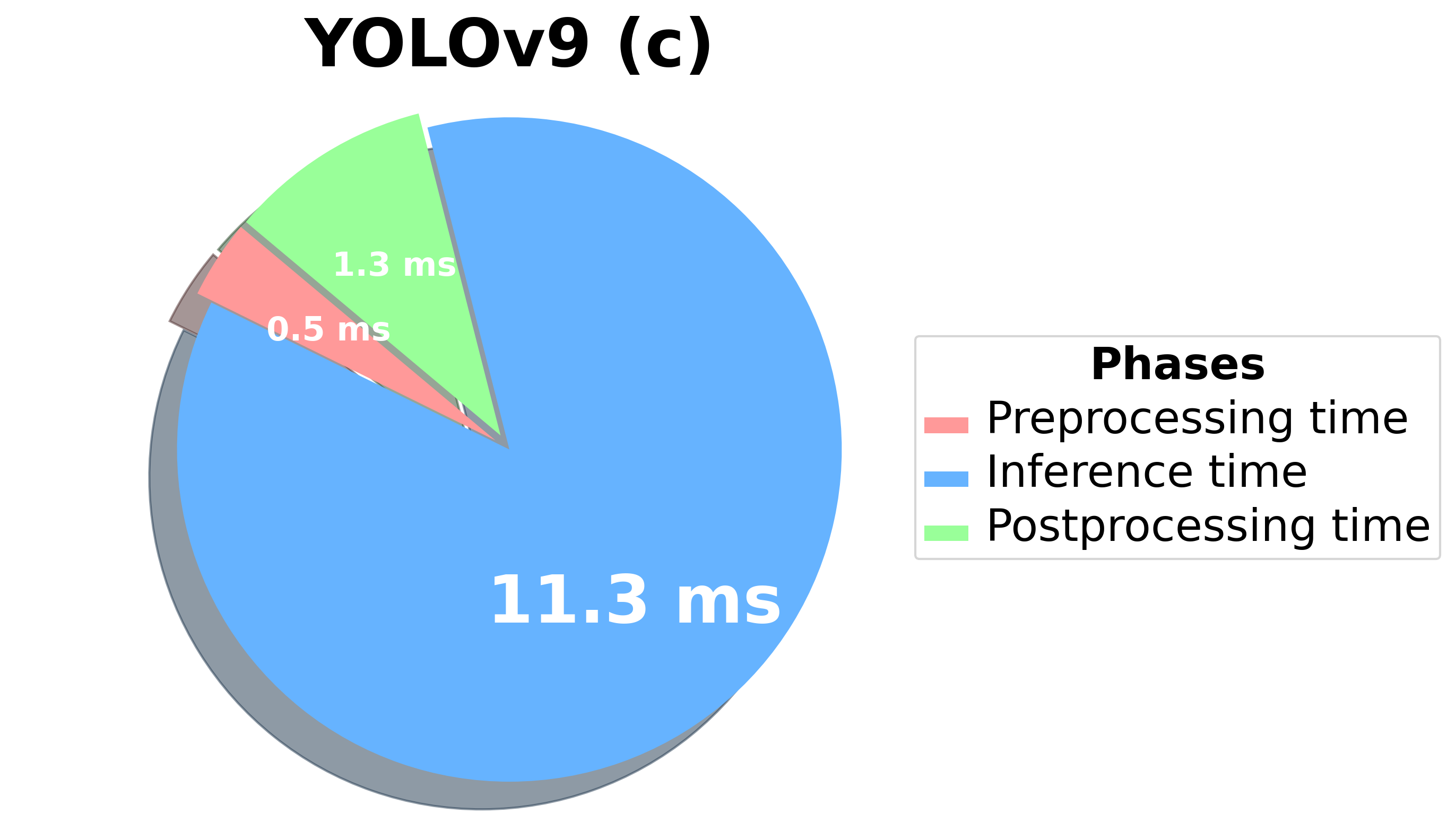}
    \label{fig:image1}
  \end{subfigure}
  \hfill
  \begin{subfigure}[b]{0.32\textwidth}
    \includegraphics[width=\textwidth]{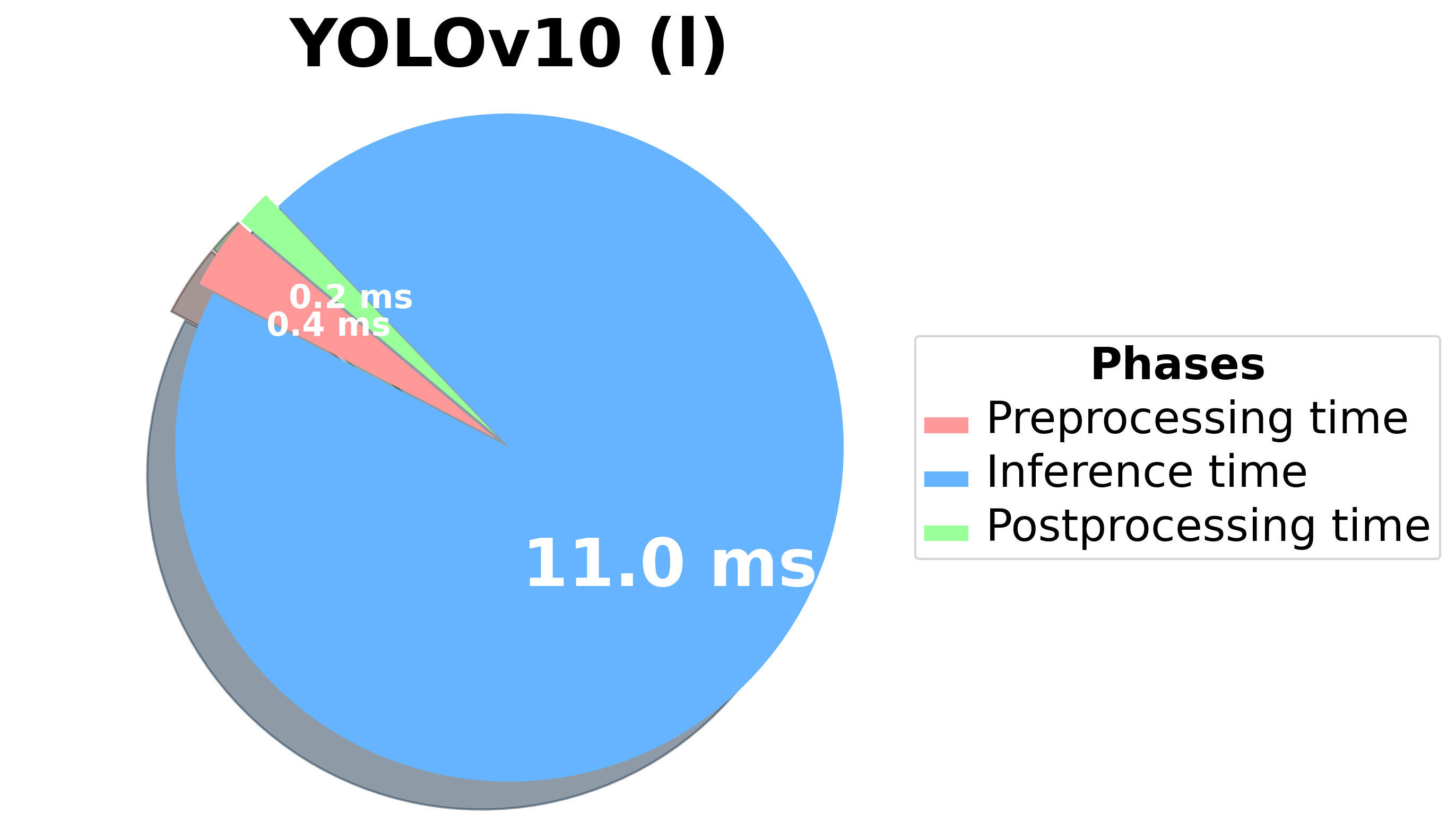}
    \label{fig:image2}
  \end{subfigure}
  \hfill
  \begin{subfigure}[b]{0.32\textwidth}
    \includegraphics[width=\textwidth]{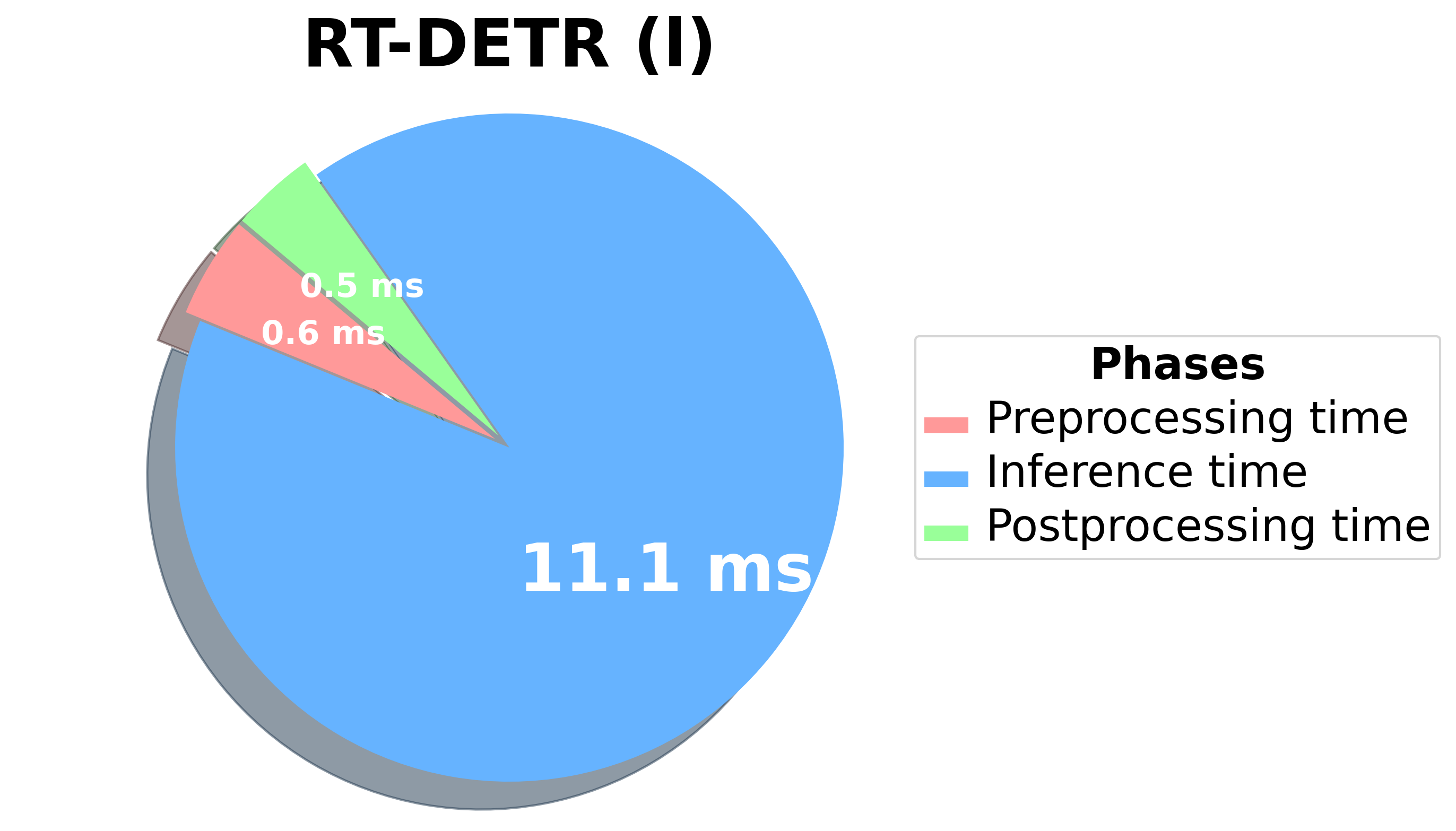}
    \label{fig:image3}
  \end{subfigure}
  \caption{Time distribution analyses for diverse models utilizing 640-pixel image resolution on NVIDIA RTX4090 GPU.}
  \label{fig:time_comparison}
\end{figure}

The analysis of GPU and CPU inference time versus mean Average Precision at mAP50 and mAP50-95 for various YOLOv9 and YOLOv10 models and RT-DETR (l) across different image sizes provides significant insights into their performance. Our results on the Intel Core i9-14900K (32-core) are shown in \Cref{fig:map50} and \Cref{fig:map50-95}. The \Cref{table:performance} presents the comprehensive results of our experimental analyzes. Additionally, \Cref{fig:new_predictions} illustrates the prediction results compared to a ground truth test image. Furthermore, the impact of NMS is evident in \Cref{fig:time_comparison}, showing that YOLOv9 has a longer post-processing time since YOLOv10 and RT-DETR do not utilize NMS.

\subsection{Inference Time vs mAP50}

For YOLOv9 models, specifically YOLOv9 (c), the mAP50 values remain high in different image sizes. For example, at an image size of 960 px, YOLOv9 (c) achieves an mAP50 of 94.0\%, with inference times of 15.9 ms on the RTX3090, 12.6 ms on the RTX4090, and 233.1 ms on the Intel CPU. As the image size decreases, the inference time also decreases, but with a slight reduction in accuracy. For instance, at 320 pixels, the mAP50 is 90.2\%, with inference times of 9.7 ms on RTX3090, 10.5 ms on RTX4090, and 38.0 ms on the Intel CPU. This trend shows that, while smaller image sizes improve inference times, they may come at the cost of reduced accuracy.

The YOLOv10 models, such as YOLOv10 (l), also demonstrate high mAP50 values with increasing image sizes. At 960 pixels, YOLOv10 (l) achieves an mAP50 of 93.7\%, with inference times of 18.2 ms on RTX3090, 11.9 ms on RTX4090, and 318.9 ms on the Intel CPU. When the image size is reduced to 320 pixels, the mAP50 drops slightly to 90.7\%, but inference times improve to 10.2 ms on RTX3090, 9.6 ms on RTX4090, and 53.0 ms on the Intel CPU.

RT-DETR (l) presents a distinct performance profile. At an image size of 960 pixels, it achieves an mAP50 of 93.0\%, with inference times of 17.8 ms on RTX3090, 12.1 ms on RTX4090, and 285.8 ms on the Intel CPU. At 320 pixels, the mAP50 decreases to 87.2\%, but the inference times are reduced to 11.3 ms on RTX3090, 9.7 ms on RTX4090, and 67.3 ms on the Intel CPU. This model generally shows longer inference times across all image sizes compared to the YOLO models, but maintains competitive mAP50 values.

\subsection{Inference Time vs mAP50-95}

When evaluating mean Average Precision at IoU thresholds ranging from 50\% to 95\% (mAP50-95), similar trends are observed. For YOLOv9 (c), at an image size of 960 pixels, the mAP50-95 is 82.2\%, with inference times of 15.9 ms on RTX3090, 12.6 ms on RTX4090, and 233.1 ms on the Intel CPU. Reducing the image size to 320 pixels results in a decrease in mAP50-95 to 73.3\%, but the inference times improve to 9.7 ms on RTX3090, 10.5 ms on RTX4090, and 38.0 ms on the Intel CPU.

YOLOv10 (l) maintains an mAP50-95 of 82.4\% at 960 pixels, with inference times of 18.2 ms on RTX3090, 11.9 ms on RTX4090, and 318.9 ms on the Intel CPU. At 320 pixels, the mAP50-95 is 72.2\%, with improved inference times of 10.2 ms on RTX3090, 9.6 ms on RTX4090, and 53.0 ms on the Intel CPU.

RT-DETR (l) achieves an mAP50-95 of 79.9\% at 960 pixels, with inference times of 17.8 ms on RTX3090, 12.1 ms on RTX4090, and 285.8 ms on the Intel CPU. At the 320-pixel resolution, the mAP50-95 decreases to 68.1\%, with inference times of 11.3 ms on RTX3090, 9.7 ms on RTX4090, and 67.3 ms on the Intel CPU.

\setlength{\tabcolsep}{4pt} 
\begin{longtable}{lcccccc}
\caption{Quantitative assessment of comprehensive model performance (mAP and inference time) across varying GPU and CPU architectures and image resolutions.}\\

\label{table:performance}\\
\toprule
      \textbf{Model} &  \textbf{mAP50-95} &  \textbf{mAP50} & \textbf{\makecell{RTX \\ 3090}} & \textbf{\makecell{RTX \\ 4090}} & \textbf{\makecell{Intel \\ CPU}} &  \textbf{\makecell{Image \\ size}} \\
\midrule
\endfirsthead
\caption[]{Quantitative assessment of comprehensive model performance (mAP and inference time) across varying GPU and CPU architectures and image resolutions.} \\
\toprule
      \textbf{Model} &  \textbf{mAP50-95} &  \textbf{mAP50} & \textbf{\makecell{RTX \\ 3090}} & \textbf{\makecell{RTX \\ 4090}} & \textbf{\makecell{Intel \\ CPU}} &  \textbf{\makecell{Image \\ size}} \\
\midrule
\endhead
\midrule
\multicolumn{7}{r}{{Continued on next page}} \\
\midrule
\endfoot

\bottomrule
\endlastfoot
YOLOv9 (c) &      82.2 &   94.0 &  15.9ms & 12.6ms  & 233.1ms  &         960 \\
YOLOv9 (c) &      81.8 &   93.7 &  13.7ms & 12.3ms  & 171.9ms  &         800 \\
YOLOv9 (c) &      80.1 &   93.4 &  12.8ms & 11.3ms  & 109.4ms  &         640 \\
YOLOv9 (c) &      79.3 &   93.2 &  10.1ms & 10.9ms  &  69.1ms  &         480 \\
YOLOv9 (c) &      73.3 &   90.2 &   9.7ms & 10.5ms  &  38.0ms  &         320 \\
YOLOv9 (m) &      81.9 &   94.0 &  15.3ms & 12.1ms  & 191.1ms  &         960 \\
YOLOv9 (m) &      81.3 &   93.9 &  12.7ms & 11.7ms  & 144.2ms  &         800 \\
YOLOv9 (m) &      79.8 &   93.4 & 11.3ms  & 11.1ms  &  90.9ms  &         640 \\
YOLOv9 (m) &      79.3 &   93.1 &  9.7ms  &  10.1ms &  58.8ms  &         480 \\
YOLOv9 (m) &      73.0 &   90.2 &  9.2ms  &   9.7ms &  34.6ms  &         320 \\
YOLOv9 (s) &      80.8 &   94.1 & 11.7ms  & 11.3ms  &  74.7ms  &         960 \\
YOLOv9 (s) &      80.4 &   94.0 & 11.4ms  & 10.7ms  &  56.8ms  &         800 \\
YOLOv9 (s) &      78.9 &   93.8 & 10.9ms  & 10.5ms  &  39.6ms  &         640 \\
YOLOv9 (s) &      77.7 &   93.5 &   9.4ms &  9.6ms  &  27.9ms  &         480 \\
YOLOv9 (s) &      70.5 &   89.2 &   9.1ms &  9.1ms  &  17.7ms  &         320 \\
YOLOv9 (t) &      80.1 &   \textbf{94.2} & 10.2ms  & 10.5ms  &  36.3ms  &         960 \\
YOLOv9 (t) &      79.4 &   94.1 &  9.8ms  & 10.3ms  &  28.8ms  &         800 \\
YOLOv9 (t) &      77.8 &   94.0 &  9.5ms  & 10.2ms  &  21.5ms  &         640 \\
YOLOv9 (t) &      75.4 &   92.6 &  9.2ms  &   9.7ms &  16.8ms  &         480 \\
YOLOv9 (t) &      67.1 &   87.4 &   9.1ms &   9.3ms &  \textbf{10.2ms}  &         320 \\
YOLOv10 (l) &      \textbf{82.4} &   93.7 &  18.2ms & 11.9ms  & 318.9ms  &         960 \\
YOLOv10 (l) &      81.7 &   93.6 &  14.8ms & 11.6ms  & 240.2ms  &         800 \\
YOLOv10 (l) &      80.2 &   93.3 & 13.6ms  & 11.0ms  & 163.1ms  &         640 \\
YOLOv10 (l) &      79.3 &   92.9 & 10.6ms  &  9.9ms  &  93.4ms  &         480 \\
YOLOv10 (l) &      72.2 &   90.7 & 10.2ms  &  9.6ms  &  53.0ms  &         320 \\
YOLOv10 (m) &      81.9 &   93.6 & 12.7ms  & 10.3ms  & 204.7ms  &         960 \\
YOLOv10 (m) &      81.6 &   93.6 & 11.3ms  & 10.1ms  & 154.9ms  &         800 \\
YOLOv10 (m) &      79.8 &   93.4 &  9.0ms  &  9.4ms  &  92.4ms  &         640 \\
YOLOv10 (m) &      79.0 &   93.3 &  8.8ms  &  9.3ms  &  57.5ms  &         480 \\
YOLOv10 (m) &      72.7 &   90.1 &  8.4ms  &  8.7ms  &  33.8ms  &         320 \\
YOLOv10 (s) &      81.4 &   93.6 &  8.1ms  &  8.4ms  & 106.3ms  &         960 \\
YOLOv10 (s) &      80.5 &   93.5 &  8.0ms  &  8.2ms  &  83.3ms  &         800 \\
YOLOv10 (s) &      79.0 &   93.1 &  7.5ms  &  8.1ms  &  59.0ms  &         640 \\
YOLOv10 (s) &      77.3 &   92.6 &  7.3ms  &   7.9ms &  31.8ms  &         480 \\
YOLOv10 (s) &      70.6 &   88.7 &  7.0ms  &  7.8ms  &  19.4ms  &         320 \\
YOLOv10 (n) &      80.5 &   93.6 &  7.5ms  &  8.0ms  &  57.2ms  &         960 \\
YOLOv10 (n) &      79.5 &   93.4 &  7.3ms  &  6.8ms  &  41.8ms  &         800 \\
YOLOv10 (n) &      77.4 &   92.4 &  7.0ms  &  6.4ms  &  31.1ms  &         640 \\
YOLOv10 (n) &      75.2 &   91.7 &   6.8ms &  5.9ms  &  19.5ms  &         480 \\
YOLOv10 (n) &      67.5 &   87.0 &  \textbf{6.2ms}  &  \textbf{5.7ms}  &  12.1ms  &         320 \\
RT-DETR (l) &      79.9 &   93.0 &  17.8ms & 12.1ms  & 285.8ms  &         960 \\
RT-DETR (l) &      78.3 &   92.9 &  14.1ms & 11.5ms  & 215.9ms  &         800 \\
RT-DETR (l) &      76.7 &   92.3 &  11.8ms & 11.1ms  & 147.3ms  &         640 \\
RT-DETR (l) &      73.8 &   90.4 &  11.5ms & 9.9ms  & 102.7ms  &         480 \\
RT-DETR (l) &      68.1 &   87.2 &  11.3ms & 9.7ms  & 67.3ms  &         320 \\

\end{longtable}

\section{Discussion}

The presented results provide a detailed comparison of the YOLOv9, YOLOv10, and RT-DETR (l) models across different image sizes and GPUs, highlighting several key findings.

First, the overall trend indicates that larger image sizes generally lead to higher mAP values for both the mAP50 and mAP50-95 metrics. This suggests that higher resolution inputs contribute to more accurate detections, though at the cost of increased inference time. This trade-off is particularly evident in the performance of the RT-DETR (l) model, which shows significant gains in mAP with larger image sizes but also exhibits slightly higher inference times compared to the YOLO models.

Secondly, the YOLOv9 and YOLOv10 models demonstrate robust performance across different image sizes and inference times. YOLOv9 (c) and YOLOv10 (l) models, in particular, consistently achieve high mAP values, indicating their suitability for applications requiring high accuracy and efficient inference times. The performance stability of the YOLOv10 models further highlights their potential for diverse application scenarios \cite{laroca2018robust}, maintaining high accuracy with minimal variation in different image sizes. Notably, YOLOv10 (n) and YOLOv10 (s) models deliver competitive accuracy with significantly lower inference times, making them ideal candidates for real-time processing where computational resources are limited.

The RT-DETR (l) model, while achieving competitive mAP values, particularly at larger image sizes, is characterized by higher inference times. This model may be more suitable for applications where accuracy is critical and real-time processing is less of a constraint. The unique performance curve of RT-DETR (l) suggests that it could be advantageous in scenarios requiring detailed object detection with high precision.

Moreover, the comparison between YOLOv9, YOLOv10, and RT-DETR (l) models underscores the importance of selecting an appropriate model based on specific application requirements. YOLO models offer a balanced trade-off between accuracy and inference time, making them versatile for both real-time and high-precision applications. In contrast, RT-DETR (l) provides superior accuracy with larger image sizes but at the expense of higher inference times. An interesting observation is the minimal difference in accuracy between image resolutions of 640 and 960 across all models. This suggests that, for many applications, utilizing an image size of 640 may be sufficient to achieve high accuracy while optimizing for faster inference times. Conversely, a noticeable accuracy drop occurs for resolutions below 640, highlighting this threshold as a crucial point in the resolution-performance trade-off.

\section{Conclusion}
\label{sec:conclusion}

This study provides a comprehensive analysis of the performance of various YOLO models (YOLOv9 and YOLOv10) and RT-DETR (l) in different image sizes and GPU \& CPU configurations with evaluation metrics including mAP50 and mAP50-95 as well as inference times. The results indicate that larger image sizes generally lead to higher mAP values, underscoring the importance of high-resolution input for accurate object detection. However, this accuracy gain comes at the cost of increased inference times, which is a critical consideration for real-time applications. YOLOv9 and YOLOv10 models demonstrate a balanced trade-off between accuracy and speed, making them suitable for a wide range of applications. The RT-DETR (l) model stands out for its high accuracy at larger image sizes, making it ideal for applications where precision is paramount and higher inference times are acceptable. Our study contributes to ongoing efforts to develop efficient and accurate object detection models, providing valuable information to select the appropriate model based on specific application requirements. 
Future research should focus on optimizing these models for various hardware configurations \cite{Nils_Herterich} and broadening their evaluation across diverse datasets, such as \cite{dang2023yoloweeds}, to improve generalizability and practical utility through cross-validation. Furthermore, the effects of implementing Slicing Aided Hyper Inference (SAHI) \cite{SAHI} in conjunction with different model compression techniques, e.g., knowledge distillation, pruning and quantization will be addressed in future work.

\bibliographystyle{unsrtnat}
\bibliography{ref}

\end{document}